\pdfoutput=1

\documentclass[11pt]{article}

\usepackage[final]{coling}

\usepackage{times}
\usepackage{latexsym}

\usepackage[T1]{fontenc}

\usepackage[utf8]{inputenc}

\usepackage{microtype}

\usepackage{inconsolata}

\usepackage{graphicx}

\usepackage{multirow}
\usepackage{float}
\title{ML-Promise: A Multilingual Dataset for Corporate Promise Verification}

\author{
  \textbf{Yohei Seki\textsuperscript{1}},
  \textbf{Hakusen Shu\textsuperscript{2}},
  \textbf{Ana\"{i}s Lhuissier\textsuperscript{3}},
  \textbf{Hanwool Lee\textsuperscript{4}},\\
  \textbf{Juyeon Kang\textsuperscript{3}},
  \textbf{Min-Yuh Day\textsuperscript{5}},
  \textbf{Chung-Chi Chen\textsuperscript{6}}
  \\
  \\
  \textsuperscript{1}Institute of Library, Information, and Media Science, University of Tsukuba, Japan,\\
  \textsuperscript{2}College of Knowledge and Library Sciences, School of Informatics, University of Tsukuba, Japan, \\
  \textsuperscript{3}3DS Outscale, France,
  \textsuperscript{4}Shinhan Securities Co., Korea,\\
  \textsuperscript{5}Graduate Institute of Information Management, National Taipei University, Taiwan
  \textsuperscript{6}AIST, Japan
}

\begin{document}
\maketitle
\begin{abstract}
Promises made by politicians, corporate leaders, and public figures have a significant impact on public perception, trust, and institutional reputation. However, the complexity and volume of such commitments, coupled with difficulties in verifying their fulfillment, necessitate innovative methods for assessing their credibility. This paper introduces the concept of Promise Verification, a systematic approach involving steps such as promise identification, evidence assessment, and the evaluation of timing for verification. We propose the first multilingual dataset, ML-Promise, which includes English, French, Chinese, Japanese, and Korean, aimed at facilitating in-depth verification of promises, particularly in the context of Environmental, Social, and Governance (ESG) reports. Given the growing emphasis on corporate environmental contributions, this dataset addresses the challenge of evaluating corporate promises, especially in light of practices like greenwashing. Our findings also explore textual and image-based baselines, with promising results from retrieval-augmented generation (RAG) approaches. This work aims to foster further discourse on the accountability of public commitments across multiple languages and domains.
\end{abstract}

\section{Introduction}
In a world where promises shape perceptions and drive decisions, the integrity of commitments made by politicians, corporate leaders, and public figures must be scrutinized. These promises, ranging from environmental sustainability to social responsibility and governance ethics, significantly influence the general public's and stakeholders' trust, as well as government and corporate reputations. Yet, the complexity and abundance of such commitments, coupled with the challenge of verifying their fulfillment, create a pressing need for innovative approaches to assess their strength and verifiability.
Recognizing the critical role of transparency and accountability in today's society, we propose a groundbreaking task: Promise Verification.

\begin{table*}[t]
  \centering
  \small
    \begin{tabular}{llrrrrr}
    \multicolumn{1}{c}{Task} & \multicolumn{1}{c}{Label} & \multicolumn{1}{c}{English} & \multicolumn{1}{c}{French} & \multicolumn{1}{c}{Chinese} & \multicolumn{1}{c}{Japanese} & \multicolumn{1}{c}{Korean} \\
    \hline
    \multirow{2}[2]{*}{Promise Identification} & Yes   & 84.5  & 80.5  & 40.2  & 74.9  & 77.5 \\
          & No    & 15.5  & 19.5  & 59.8  & 25.1  & 22.5 \\
    \hline
    \multirow{2}[2]{*}{Supporting Evidence} & Yes   & 20.1  & 71.6  & 20.1  & 66.4  & 75.6 \\
          & No    & 79.9  & 28.4  & 79.9  & 33.6  & 24.4 \\
    \hline
    \multirow{3}[2]{*}{Clarity of Promise-Evidence Pair} & Clear & 53.3  & 56.6  & 64.7  & 61.2  & 94.8 \\
          & Not Clear & 42.9  & 41.9  & 35.3  & 34.7  & 5.2 \\
          & Misleading & 3.8   & 1.5   & 0.0   & 4.1   & 0.0 \\
    \hline
    \multirow{4}[1]{*}{Timing for Verification} & Within 2 years & 1.9   & 12.4  & 37.5  & 7.3   & 45.5 \\
          & 2-5 years & 14.1  & 15.0  & 10.0  & 9.3   & 8.4 \\
          & Longer than 5 years & 9.0   & 21.6  & 15.0  & 18.7  & 17.5 \\
          & Other & 75.0  & 51.0  & 37.5  & 64.7  & 28.7 \\
    \end{tabular}%
  \caption{Label distribution in each language. (\%)}
  \label{tab:distribution}%
\end{table*}%

To perform promise verification, several steps are required, including (1) identifying the promise, (2) linking the promise with supporting evidence, (3) assessing the clarity of the promise-evidence pair, and (4) inferring the timing for verifying the promise. For example, after identifying a promise, the availability of evidence to support that the speaker is taking action to fulfill the promise could serve as a coarse-grained evaluation of the promise's quality. However, the clarity of the evidence may also affect the quality of the promise, which would be a fine-grained evaluation. Additionally, whether the speaker provides a clear timeline for verifying the promise is an important criterion. For instance, ``we will achieve net zero carbon emissions within five years'' is a stronger promise than ``we will achieve net zero carbon emissions.'' Following this line of thought, this paper proposes the first multilingual dataset for in-depth promise verification, including Chinese, English, French, Japanese, and Korean.

In recent years, increasing emphasis has been placed on companies’ environmental contributions, especially in addressing climate change, deforestation, and compliance with labor conditions and governance, when evaluating their investment value. In the evolving landscape of ESG (environmental, social, and governance) criteria, the ability to accurately assess a company's promises and adherence to its ESG promises has become paramount. However, unlike traditional financial statements, ESG reports still lack clear standards regarding corporate promises. This allows some companies to use misleading information to project an overly positive environmental image, a practice known as greenwashing. As \citet{nina_gorovaia__2024} points out, companies involved in environmental misconduct tend to produce longer, more positive, and more frequent reports. We hypothesize that such reports may lack substantive evidence, or the information presented may be irrelevant or ambiguous, leading to misinterpretation.
To this end, the proposed dataset, ML-Promise, focuses on ESG reports released by corporations in five countries: the U.K., France, Taiwan, Japan, and Korea.

In addition to exploring text-based baselines, we also provide pilot results on image-based experiments, as most reports are released in PDF format. Our experiment further shows that the retrieval-augmented generation (RAG) approach~\cite{lewis2020} can help in some language cases. Although we do not find a silver bullet for all languages and tasks, we hope the proposed dataset can open a new chapter in discussions on the responsibility of promises, especially those from public figures.

\section{Related Work}
\label{sec:related_work}

Recent studies have sought to improve the analysis of ESG or sustainability reports for estimating company values using contextual embedding approaches. For example, \citet{Gutierrez-Bustamante2022Natural} evaluated sustainability reports from publicly listed companies in Nordic countries using latent semantic analysis (LSA) and the global vectors for word representation (GloVe) model, enhancing document retrieval performance based on similarity. \citet{garigliotti-2024-sdg} explored the integration of sustainable development goals (SDGs) into environmental impact assessments (EIAs) using a RAG framework powered by large language models (LLMs). Their work focused on two tasks: detecting SDG targets within EIA reports and identifying relevant textual evidence, specifically in European contexts. \citet{hillebrand2023} introduced sustain.AI, a context-aware recommender system designed to analyze sustainability reports in response to increasing corporate social responsibility (CSR) regulations. The system, based on a BERT architecture, identified relevant sections of lengthy reports using global reporting initiative (GRI) indicators and demonstrated strong performance on datasets from German companies. Previous studies have a few shortcomings. First, most of them focus solely on reports from one country. Second, none of them attempt to analyze corporate promises, despite the abundance of sustainability reports. To address these problems, our study extends these works by focusing on multilingual companies from both European and Asian regions, including Taiwan, the UK, France, Japan, and Korea. With the proposed new task, we aim to highlight the importance of anti-greenwashing by evaluating corporate promises in ESG reports.

\begin{table*}[t]
  \centering
  \small
    \begin{tabular}{llrrrrr}
       Approach   & \multicolumn{1}{c}{Task} & \multicolumn{1}{c}{English} & \multicolumn{1}{c}{French} & \multicolumn{1}{c}{Chinese} & \multicolumn{1}{c}{Japanese} & \multicolumn{1}{c}{Korean} \\
    \hline
    \multirow{4}[2]{*}{w/o RAG} & Promise Identification (PI) & 0.842  & \textbf{0.816} & 0.521  & \textbf{0.670} & \textbf{0.849} \\
          & Supporting Evidence (SE) & 0.680  & \textbf{0.746} & 0.163  & 0.720  & \textbf{0.792} \\
          & Clarity of Promise-Evidence Pair (CPEP) & 0.411  & 0.443  & 0.569  & 0.450  & 0.897  \\
          & Timing for Verification (TV) & 0.636  & 0.523  & 0.317  & 0.632  & 0.406  \\
    \hline
    \multirow{4}[1]{*}{w/ RAG} & Promise Identification & \textbf{0.866} & 0.798  & \textbf{0.540} & 0.659  & 0.807  \\
          & Supporting Evidence & \textbf{0.757} & 0.732  & \textbf{0.503} & \textbf{0.850} & 0.774  \\
          & Clarity of Promise-Evidence Pair & \textbf{0.467} & \textbf{0.487} & \textbf{0.628} & \textbf{0.465} & \textbf{0.939} \\
          & Timing for Verification & \textbf{0.693} & \textbf{0.601} & \textbf{0.469} & \textbf{0.684} & \textbf{0.571} \\
    \end{tabular}%
  \caption{Experimental Results (F1-Score). The best performance in each language is denoted in \textbf{bold}.}
  \label{tab:text_results}
\end{table*}%

\section{ML-Promise}
\subsection{Task Design}

We collect ESG reports from five countries: the UK, France, Taiwan, Japan, and Korea. The annotators are native speakers of the target language or are familiar with the language at the work level. The task designs are as follows when given a paragraph in the ESG reports.

\begin{enumerate}
    \small
    \item \textbf{Promise Identification}: This is a boolean label (Yes/No) based on whether a promise exists.
    
    \item \textbf{Supporting Evidence}: This is a boolean label (Yes/No) based on whether supporting evidence exists.
    
    \item \textbf{Clarity of the Promise-Evidence Pair}: We designed three labels (Clear/Not Clear/Misleading) for this task, which should depend on the clarity of the given evidence in relation to the promise.
    
    \item \textbf{Timing for Verification}: Following the MSCI guidelines and previous work~\cite{tseng2023dynamicesg}, we set timing labels (within 2 years/2-5 years/longer than 5 years/other) to indicate when readers/investors should return to verify the promise. Here, ``other'' denotes the promise has already been verified or doesn't have a specific timing to verify it.

\end{enumerate}

\subsection{Statistics}
Finally, we obtained 3,010 instances, i.e., 600 for each language and 10 additional instances in the Chinese dataset. 
The Cohen's $\kappa$ agreement~\cite{cohen1960,mchugh2012interrater} for these tasks is approximately 0.65-0.96, 0.71-0.88, 0.62-0.80, and 0.60-0.89, respectively.
Table~\ref{tab:distribution} presents the distribution of the proposed ML-Promise dataset. First, we observe that around 35-40\% of the evidence is ``not clear'' in supporting the associated promises in four out of five languages. This highlights the necessity of the proposed task for evaluating the quality of the promise-evidence pairs from corporations. Furthermore, about 4\% of instances contain (potentially) misleading evidence in the English and Japanese datasets. It is crucial for corporations to re-examine this evidence, and it is also essential for supervisory authorities to monitor these instances. Second, we noted that corporations in Taiwan and Korea tend to make more short-term promises (within 2 years), whereas corporations in the remaining countries tend to make longer-term promises. This finding shows the need for a multilingual comparison of ESG reports across different countries, as the narrative styles vary among them.

\section{Experiment}
\subsection{Methods}
RAG~\cite{lewis2020} was introduced as a method to enhance LLMs by integrating external knowledge sources. This approach combines retrieval mechanisms with generative models, producing more accurate and contextually relevant outputs. \citet{honggang_yu__2024} highlights the advantages of RAG systems, particularly their ability to extract domain-specific information. By incorporating external retrieval processes, RAG enables generative models to access a broader, field-specific knowledge base, improving the accuracy and relevance of responses. This capability is especially important for handling domain-specific queries, an area where existing LLMs often encounter difficulties. \citet{yujuan_ding__2024} discusses training strategies for RAG, including independent, sequential, and joint methods, which can be tailored to optimize retrieval and generation for specific domains. For Chinese language applications, \citet{shuting_wang__2024} emphasizes the importance of domain-specific corpora over general knowledge sources. 
\citet{ozgur_ardic__2024} applied RAG to analyze sustainability reports from ten Turkish companies, focusing on ESG factors. The study found BM25 outperformed BERTurk in retrieving relevant sections, highlighting the effectiveness of advanced retrieval techniques.

Following the findings of previous studies, we also explore and design the RAG approach for the proposed tasks. Specifically, when given a paragraph, we first retrieve the six most similar samples in the training set. We leveraged Multilingual E5 Text Embeddings~\cite{wang2024multilingual} to calculate the cosine similarity between target paragraphs and instances from the training set. Then, we provide the top-six examples for the LLM to perform in-context learning~\cite{dong2022survey}. In our experiment, we use GPT-4o as the base LLM.

\begin{table}[t]
  \centering
    \resizebox{\columnwidth}{!}{
    \begin{tabular}{ll|rr|rr}
        \multicolumn{1}{c}{\multirow{2}[1]{*}{RAG}}  & \multicolumn{1}{c|}{\multirow{2}[1]{*}{Task}} & \multicolumn{2}{c|}{Chinese} & \multicolumn{2}{c}{Korean} \\
          &       & \multicolumn{1}{c}{Image-Based} & \multicolumn{1}{c|}{Text-Based} & \multicolumn{1}{c}{Image-Based} & \multicolumn{1}{c}{Text-Based} \\
    \hline
    \multirow{4}[2]{*}{w/o} & PI & 0.530  & 0.521  & 0.837  & \textbf{0.849} \\
          & SE & 0.124  & 0.163  & 0.812  & 0.792  \\
          & CPEP & 0.510  & 0.569  & 0.922  & 0.897  \\
          & TV & 0.202  & 0.317  & 0.201  & 0.406  \\
    \hline
    \multirow{4}[1]{*}{w/} & PI & \textbf{\underline{0.580}} & \underline{0.540}  & \underline{0.843}  & 0.807  \\
          & SE & \textbf{\underline{0.512}} & \underline{0.503}  & \textbf{\underline{0.845}} & 0.774  \\
          & CPEP & \underline{0.618}  & \textbf{\underline{0.628}} & 0.893  & \textbf{\underline{0.939}} \\
          & TV & \underline{0.297}  & \textbf{\underline{0.469}} & \underline{0.330}  & \textbf{\underline{0.571}} \\
    \end{tabular}%
    }
  \caption{Image-based experimental results. \textbf{Bolded} denotes the best performance in each language. \underline{Underlined} denotes performance with RAG better than that without RAG.}
  \label{tab:Image-based experimental results}%
\end{table}%

\subsection{Experimental Results}
In the experiment, we randomly select 200 instances from each language as the test set, and the remaining instances are used for training.
We use the F1 score to evaluate the performance of each task. Table~\ref{tab:text_results} shows the performance of each task in each language. First, the performance of most tasks improves when adopting RAG. Specifically, for English and Chinese, all tasks perform better when using RAG. Second, RAG enhances performance in estimating the clarity of the promise-evidence pair and inferring the timing for verification, regardless of the language used. These results suggest the usefulness of RAG in these two novel tasks. Additionally, the findings demonstrate the value of the proposed annotations. With the proposed dataset, the performance of fine-grained promise evaluation can be improved. Third, although the performance in promise identification and supporting evidence identification tasks may slightly decrease in French, Japanese, and Korean, the declines are minimal (less than 2\% in most cases). 
These results suggest that the method for retrieving and suggesting samples similar to the paragraph requires refinement for imbalanced boolean datasets. In future work, we will focus on improving the RAG approach by extracting balanced samples, particularly for minor labels.

\section{Discussion}
\subsection{Image-Based Experiment}
We noticed a significant difference between Taiwan/Korea reports and the reports from other countries.\footnote{We provide some examples in Appendix~\ref{sec:Report Examples}.} The reports from these two countries utilize a large number of graphs instead of textual descriptions. This observation raises the question of whether we could use multimodal LLMs to read PDF files directly instead of relying on extracted text. To explore this, we align the annotations with a PDF page and employ GPT-4o to reassess the tasks using an image as input. 
For RAG, we leveraged E5-V Universal Embeddings~\cite{jiang2024e5vuniversalembeddingsmultimodal} to calculate the cosine similarity between target pages and instances from the training set.

Table~\ref{tab:Image-based experimental results} presents the performance. First, using GPT-4o with image input reduces performance in three out of four tasks in the Chinese dataset and in two out of four tasks in the Korean dataset. Second, RAG improves the performance of most tasks when using image input. Third, with RAG, the performance in promise identification and supporting evidence identification tasks improves with Chinese image input, and the performance of supporting evidence identification improves with Korean image input. However, for estimating the clarity of the promise-evidence pair and inferring the timing for verification, using text input with RAG remains superior. In summary, our experimental results suggest that image input should be used for PI and SE tasks, while text input is preferable for CPEP and TV tasks. Additionally, RAG performs well regardless of input type.

\begin{table}[t]
  \centering
  \small
    \begin{tabular}{lll|r}
    Input & \multicolumn{1}{c}{RAG} & \multicolumn{1}{c|}{Task} & \multicolumn{1}{c}{ROUGE-L} \\
    \hline
    \multirow{4}[4]{*}{Text} & \multirow{2}[2]{*}{w/o} & Promise Extraction    & 0.012 \\
          &       & Evidence Extraction    & 0.007 \\
\cline{2-4}          & \multirow{2}[2]{*}{w/} & Promise Extraction    & 0.101 \\
          &       & Evidence Extraction    & 0.139 \\
    \hline
    \multirow{4}[3]{*}{Image} & \multirow{2}[2]{*}{w/o} & Promise Extraction    & 0.190 \\
          &       & Evidence Extraction    & 0.230 \\
\cline{2-4}          & \multirow{2}[1]{*}{w/} & Promise Extraction    & \textbf{0.240} \\
          &       & Evidence Extraction    & \textbf{0.317} \\
    \end{tabular}%
  \caption{Results of promise and evidence extraction.}
  \label{tab:Results of promise and evidence extraction}%
\end{table}%

\subsection{Promise and Evidence Extraction}
In the previous section, we explored the promise and evidence identification tasks. However, the task can also be formulated in an extractive manner. Instead of only outputting a yes or no, 
we can also ask models to extract the promise and evidence from the report. We provide additional annotations in the Chinese dataset and experiment in multimodal settings with and without RAG. The ROUGE-L~\cite{lin-2004-rouge} score is used to evaluate extraction performance. Table~\ref{tab:Results of promise and evidence extraction} presents the results. These results indicate that the best performance is achieved in the image-based setting with RAG for both promise and evidence extraction. This emphasizes the importance of exploring multimodal input for ESG report understanding.

\section{Conclusion}
In this paper, we introduce the concept of Promise Verification, a novel task aimed at evaluating the credibility and fulfillment of promises made by corporations, particularly in the context of ESG reports. We propose the first multilingual dataset, ML-Promise, to emphasize the importance of assessing corporate environmental and social promises. Our results demonstrate that RAG improves performance, while also showing the potential of multimodal approaches in promise verification. Our annotations will be released under the CCBY-NC-SA 4.0 license. We hope this work serves as a foundation for the robustness of promise verification systems and contributes to greater accountability in corporate and public disclosures.

\section*{Limitation}
Several limitations warrant discussion. First, although the ML-Promise dataset includes five languages?Chinese, English, French, Japanese, and Korean?its scope is still limited to a few countries and may not fully capture the diversity of corporate promise communication styles globally. The dataset focuses on ESG reports from specific regions, which may limit the generalizability of the findings to other languages and cultural contexts. Future studies can follow our design to expand the dataset to include more regions and languages, which could enhance the robustness and applicability of the proposed methods.
Second, although the study uses RAG to improve performance, the results show that this approach does not consistently outperform baseline models across all languages and tasks. These inconsistencies suggest that RAG may require further optimization or task-specific adjustments, particularly in handling the nuances of each language and dataset structure.

These limitations and our findings highlight areas for future research, including expanding the dataset, refining the RAG approach, enhancing multimodal learning, and addressing the inherent ambiguities in corporate ESG reporting.

\bibliography{coling_latex}

\begin{thebibliography}{16}
\providecommand{\natexlab}[1]{#1}

\bibitem[{Ardic et~al.(2024)Ardic, Ozturk, Demirtas, and
  Arslan}]{ozgur_ardic__2024}
Ozgur Ardic, Mahiye~Uluyagmur Ozturk, Irem Demirtas, and Secil Arslan. 2024.
\newblock \href {https://doi.org/10.1109/SIU61531.2024.10600994} {{Information
  Extraction from Sustainability Reports in Turkish through RAG Approach}}.
\newblock In \emph{2024 32nd Signal Processing and Communications Applications
  Conference (SIU)}, pages 1--4.

\bibitem[{Cohen(1960)}]{cohen1960}
Jacob Cohen. 1960.
\newblock \href {https://doi.org/10.1177/001316446002000104} {{A Coefficient of
  Agreement for Nominal Scales}}.
\newblock \emph{Educational and Psychological Measurement}, 20(1):37--46.

\bibitem[{Dong et~al.(2022)Dong, Li, Dai, Zheng, Wu, Chang, Sun, Xu, and
  Sui}]{dong2022survey}
Qingxiu Dong, Lei Li, Damai Dai, Ce~Zheng, Zhiyong Wu, Baobao Chang, Xu~Sun,
  Jingjing Xu, and Zhifang Sui. 2022.
\newblock \href {https://arxiv.org/abs/2301.00234} {{A Survey on In-context
  Learning}}.
\newblock \emph{arXiv preprint arXiv:2301.00234}.

\bibitem[{Fan et~al.(2024)Fan, Ding, Ning, Wang, Li, Yin, Chua, and
  Li}]{yujuan_ding__2024}
Wenqi Fan, Yujuan Ding, Liangbo Ning, Shijie Wang, Hengyun Li, Dawei Yin,
  Tat-Seng Chua, and Qing Li. 2024.
\newblock \href {https://arxiv.org/abs/2405.06211} {{A Survey on RAG Meeting
  LLMs: Towards Retrieval-Augmented Large Language Models}}.
\newblock \emph{Preprint}, arXiv:2405.06211.

\bibitem[{Garigliotti(2024)}]{garigliotti-2024-sdg}
Dario Garigliotti. 2024.
\newblock \href {https://aclanthology.org/2024.climatenlp-1.19} {{SDG target
  detection in environmental reports using Retrieval-augmented Generation with
  LLMs}}.
\newblock In \emph{Proceedings of the 1st Workshop on Natural Language
  Processing Meets Climate Change (ClimateNLP 2024)}, pages 241--250, Bangkok,
  Thailand. Association for Computational Linguistics.

\bibitem[{Gorovaia and Makrominas(2024)}]{nina_gorovaia__2024}
Nina Gorovaia and Michalis Makrominas. 2024.
\newblock \href {https://doi.org/10.1111/eufm.12509} {{Identifying greenwashing
  in corporate-social responsibility reports using natural-language
  processing}}.
\newblock \emph{European Financial Management}.

\bibitem[{Gutierrez-Bustamante and
  Espinosa-Leal(2022)}]{Gutierrez-Bustamante2022Natural}
Marcelo Gutierrez-Bustamante and Leonardo Espinosa-Leal. 2022.
\newblock \href {https://doi.org/10.3390/su14159165} {{Natural Language
  Processing Methods for Scoring Sustainability Reports? A Study of Nordic
  Listed Companies}}.
\newblock \emph{Sustainability}, 14(15).

\bibitem[{Hillebrand et~al.(2023)Hillebrand, Pielka, Leonhard, Deu\ss{}er,
  Dilmaghani, Kliem, Loitz, Morad, Temath, Bell, Stenzel, and
  Sifa}]{hillebrand2023}
Lars Hillebrand, Maren Pielka, David Leonhard, Tobias Deu\ss{}er, Tim
  Dilmaghani, Bernd Kliem, R\"{u}diger Loitz, Milad Morad, Christian Temath,
  Thiago Bell, Robin Stenzel, and Rafet Sifa. 2023.
\newblock \href {https://doi.org/10.1145/3594536.3595131} {{sustain.AI: a
  Recommender System to analyze Sustainability Reports}}.
\newblock In \emph{Proceedings of the Nineteenth International Conference on
  Artificial Intelligence and Law}, ICAIL '23, page 412?416, New York, NY, USA.
  Association for Computing Machinery.

\bibitem[{Jiang et~al.(2024)Jiang, Song, Zhang, Huang, Deng, Sun, Zhang, Wang,
  and Zhuang}]{jiang2024e5vuniversalembeddingsmultimodal}
Ting Jiang, Minghui Song, Zihan Zhang, Haizhen Huang, Weiwei Deng, Feng Sun,
  Qi~Zhang, Deqing Wang, and Fuzhen Zhuang. 2024.
\newblock \href {https://arxiv.org/abs/2407.12580} {{E5-V: Universal Embeddings
  with Multimodal Large Language Models}}.
\newblock \emph{Preprint}, arXiv:2407.12580.

\bibitem[{Lewis et~al.(2020)Lewis, Perez, Piktus, Petroni, Karpukhin, Goyal,
  K\"{u}ttler, Lewis, Yih, Rockt\"{a}schel, Riedel, and Kiela}]{lewis2020}
Patrick Lewis, Ethan Perez, Aleksandra Piktus, Fabio Petroni, Vladimir
  Karpukhin, Naman Goyal, Heinrich K\"{u}ttler, Mike Lewis, Wen-tau Yih, Tim
  Rockt\"{a}schel, Sebastian Riedel, and Douwe Kiela. 2020.
\newblock \href {https://dl.acm.org/doi/abs/10.5555/3495724.3496517}
  {{Retrieval-augmented generation for knowledge-intensive NLP tasks}}.
\newblock In \emph{Proceedings of the 34th International Conference on Neural
  Information Processing Systems}, Red Hook, NY, USA. Curran Associates Inc.

\bibitem[{Lin(2004)}]{lin-2004-rouge}
Chin-Yew Lin. 2004.
\newblock \href {https://aclanthology.org/W04-1013} {{ROUGE}: A package for
  automatic evaluation of summaries}.
\newblock In \emph{Text Summarization Branches Out}, pages 74--81, Barcelona,
  Spain. Association for Computational Linguistics.

\bibitem[{McHugh(2012)}]{mchugh2012interrater}
Mary~L McHugh. 2012.
\newblock Interrater reliability: the kappa statistic.
\newblock \emph{Biochemia medica}, 22(3):276--282.

\bibitem[{Tseng et~al.(2023)Tseng, Chen, Huang, and Chen}]{tseng2023dynamicesg}
Yu-Min Tseng, Chung-Chi Chen, Hen-Hsen Huang, and Hsin-Hsi Chen. 2023.
\newblock \href {https://doi.org/10.1145/3583780.3615118} {{DynamicESG: A
  Dataset for Dynamically Unearthing ESG Ratings from News Articles}}.
\newblock In \emph{Proceedings of the 32nd ACM International Conference on
  Information and Knowledge Management}, pages 5412--5416.

\bibitem[{Wang et~al.(2024{\natexlab{a}})Wang, Yang, Huang, Yang, Majumder, and
  Wei}]{wang2024multilingual}
Liang Wang, Nan Yang, Xiaolong Huang, Linjun Yang, Rangan Majumder, and Furu
  Wei. 2024{\natexlab{a}}.
\newblock \href {https://arxiv.org/abs/2402.05672} {{Multilingual E5 Text
  Embeddings: A Technical Report}}.
\newblock \emph{arXiv preprint arXiv:2402.05672}.

\bibitem[{Wang et~al.(2024{\natexlab{b}})Wang, Liu, Song, Cheng, Fu, Guo, Fang,
  Zhu, and Dou}]{shuting_wang__2024}
Shuting Wang, Jiongnan Liu, Shiren Song, Jiehan Cheng, Yuqi Fu, Peidong Guo,
  Kun Fang, Yutao Zhu, and Zhicheng Dou. 2024{\natexlab{b}}.
\newblock \href {https://arxiv.org/abs/2406.05654} {{DomainRAG: A Chinese
  Benchmark for Evaluating Domain-specific Retrieval-Augmented Generation}}.
\newblock \emph{Preprint}, arXiv:2406.05654.

\bibitem[{Yu et~al.(2024)Yu, Gan, Zhang, Tong, Liu, and
  Liu}]{honggang_yu__2024}
Hao Yu, Aoran Gan, Kai Zhang, Shiwei Tong, Qi~Liu, and Zhaofeng Liu. 2024.
\newblock \href {https://arxiv.org/abs/2405.07437} {{Evaluation of
  Retrieval-Augmented Generation: A Survey}}.
\newblock \emph{Preprint}, arXiv:2405.07437.

\end{thebibliography}

\appendix
\section{Report Examples}
\label{sec:Report Examples}
We provide five ESG report examples in this section, and please refer to our training set for more instances: \url{https://drive.google.com/drive/folders/1wWwo5DBY2qFj2KSEqjkjinuK5CB5ku5K?usp=sharing}

\begin{itemize}

    \item English example: \url{https://www.burberryplc.com/content/dam/burberry/corporate/oar/2021/pdf/Burberry_2020-21_ESG.pdf}

    \item French example: \url{https://www.remy-cointreau.com/app/uploads/2024/04/REMY_COINTREAU_RAPPORT_RSE_2023.pdf}
    
    \item Chinese example: \url{https://esg.tsmc.com/zh-Hant/file/public/c-all_111.pdf}

    \item Japanese example:    \url{https://global.honda/jp/sustainability/report/pdf/2023/Honda-SR-2023-jp-004.pdf}

    \item Korean example: \url{https://kind.krx.co.kr/external/2024/07/24/000126/20240724000069/2023%20POSCO%20Holdings%20Sustainability%20Report%20%28KOR%29.pdf}

\end{itemize}

\end{document}